%% file: main.tex
\documentclass[letterpaper, 10 pt, conference]{ieeeconf}  

\IEEEoverridecommandlockouts                              

\usepackage{amssymb}
\usepackage{times}
\makeatletter
\let\NAT@parse\undefined
\makeatother
\usepackage[numbers,sort&compress]{natbib} 

\usepackage{multicol}

\usepackage[bookmarks=true]{hyperref}
\usepackage{amsmath}
\usepackage{graphicx}
\usepackage{cuted}
\usepackage{caption}
\usepackage{multirow}
\usepackage[table,xcdraw]{xcolor}
\usepackage{dsfont}
\usepackage{subcaption}
\usepackage{wrapfig}
\let\labelindent\relax
\usepackage{enumitem}
\setlist[description]{leftmargin=0.6cm, itemsep=2pt, topsep=2pt}
\usepackage[linesnumbered]{algorithm2e}
\RestyleAlgo{ruled}

\usepackage{amsthm}

\newtheorem{mydef}{Definition}

\newcommand{\ve}[1]{\boldsymbol{#1}}
\newcommand{\Tau}{\mathrm{T}}

\newcommand{\mosaic}{\textsc{Mosaic}}

\renewcommand{\baselinestretch}{0.965}

\title{\LARGE \bf
\mosaic{}: Skill-Centric Manipulation Planning with Physics Simulation
}

\author{
Itamar Mishani, Yorai Shaoul, and Maxim Likhachev \\
Robotics Institute, School of Computer Science\\
Carnegie Mellon University, 
United States\\
\texttt{\{imishani, yoraish, maxim\}@cs.cmu.edu}
}

\begin{document}

\maketitle


\begin{strip}       
    \centering
    \vspace{-40pt}
    \begin{minipage}{\linewidth}
        \centering
        \includegraphics[width=0.9\textwidth]{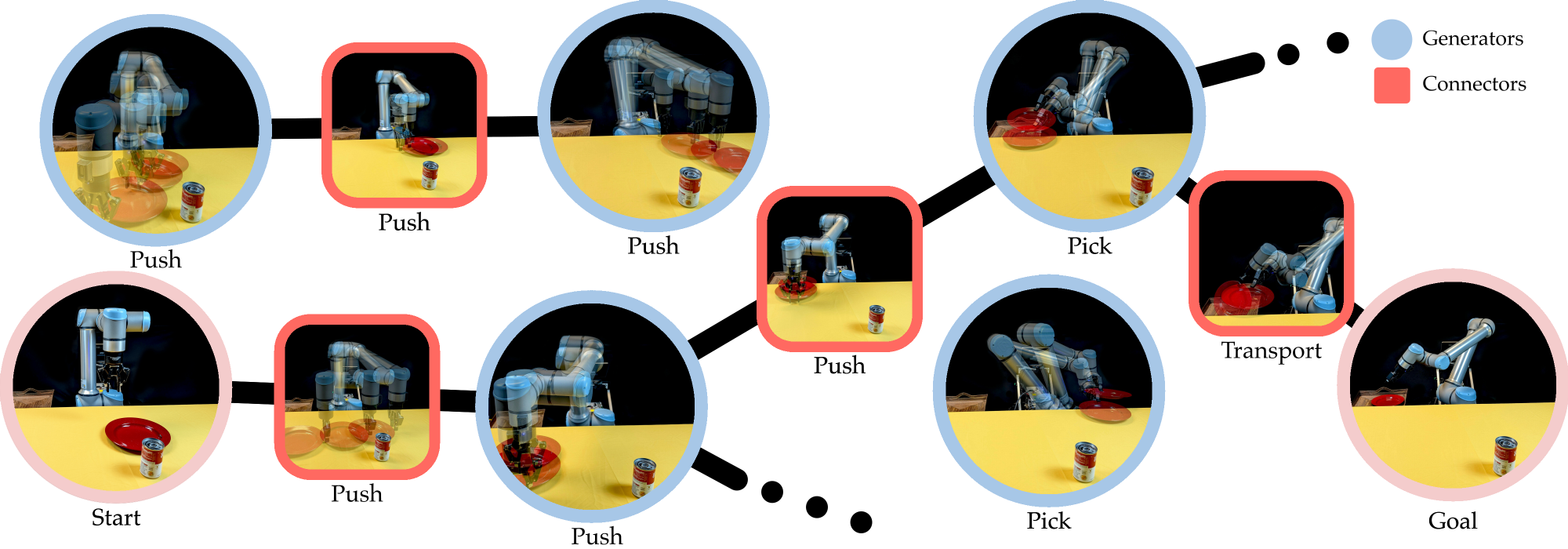}
    \end{minipage}
    
    \vspace{2.5pt}
    
    \begin{minipage}{\linewidth}
        \centering
    \end{minipage}
    \captionof{figure}{
    \mosaic{} solves long-horizon manipulation tasks by generating local skill trajectories (circles) and connecting those with connector skills (squares). \mosaic{} capitalizes on the skills themselves to guide the exploration process toward regions where they are likely to succeed -- enabling effective composition of generic local skills to solve complex tasks.
    }
    \label{fig:moto} 
\end{strip}


\input{tex/abstract}



\input{tex/introduction}


\input{tex/related_work}

\input{tex/problem_formulation}


\input{tex/mosaic}


\input{tex/experimental_analysis}


\input{tex/conclusion}




\bibliographystyle{IEEEtranN}
\bibliography{main}

\input{tex/appendix}

\end{document}

%% file: tex/abstract.tex
\begin{abstract}

Planning long-horizon manipulation motions using a set of predefined skills is a central challenge in robotics; solving it efficiently could enable general-purpose robots to tackle novel tasks by flexibly composing generic skills.
Solutions to this problem lie in an infinitely vast space of parameterized skill sequences -- a space where common incremental methods struggle to find sequences that have non-obvious intermediate steps. Some approaches reason over lower-dimensional, symbolic spaces, which are more tractable to explore but may be brittle and are laborious to construct.
In this work, we introduce \mosaic{}, a skill-centric, multi-directional planning approach that targets these challenges by reasoning about which skills to employ and where they are most likely to succeed, by utilizing physics simulation to estimate skill execution outcomes.
Specifically, \mosaic{} employs two complementary skill families: \textit{Generators}, which identify ``islands of competence'' where skills are demonstrably effective, and \textit{Connectors}, which link these skill-trajectories by solving boundary value problems. 
By focusing planning efforts on regions of high competence, \mosaic{} efficiently discovers physically-grounded solutions. 
We demonstrate its efficacy on complex long-horizon problems in both simulation and the real world, using a diverse set of skills including generative diffusion models, motion planning algorithms, and manipulation-specific models. 
Visit \href{https://skill-mosaic.github.io}{skill-mosaic.github.io} for demonstrations and examples.

\end{abstract}

%% file: tex/introduction.tex
\section{introduction}
\label{sec:introduction}
Recent progress in robotics has enabled learning complex manipulation skills that can be executed reliably under certain controlled scenarios. 
However, deploying robots in unstructured environments such as homes and offices requires solving longer-horizon tasks using skills not necessarily trained for those conditions. 
For example, a robot tidying a cluttered table must autonomously discover how to reorient plates for stable grasps, avoid wet areas, and coordinate sweeping motions. 
Such scenarios demand reasoning over long horizons and composing a diverse set of imperfect skills--not only sequencing them but also identifying conditions under which they are likely to succeed. 

Existing approaches typically represent two extremes. 
On the one hand, policy learning methods succeed in short-horizon tasks but struggle with long-horizon reasoning and skill composition due to large data requirements and poor generalization without substantial retraining. 
On the other hand, Task and Motion Planning (TAMP) frameworks offer hierarchical reasoning but are constrained by their reliance on explicit symbolic representations, which often fail to capture critical geometric and physical details, compromising robustness in open-world settings. 
Surprisingly, despite their varying strategies, many existing algorithms are fundamentally similar in that they adopt a goal-directed search, exploring outward from the start state or backward from the goal. However, this approach can be inefficient when solutions depend on non-obvious intermediate steps that do not offer an immediate indication of progress toward the goal.

In this work, we introduce \mosaic{}, an algorithmic framework for long-horizon planning that positions skills as key stakeholders in the planning process, enabled by a high-fidelity physics simulator that evaluates their feasibility during search. Rather than being constrained by directed search, \mosaic{} explores multiple directions simultaneously, anchoring its search in regions of the state space where skills are most likely to succeed. The approach leverages \emph{generator skills} to propose effective skill-trajectories and world configurations, while utilizing \emph{connector skills} to generate transitions between these regions by solving boundary value problems. Using simulation as world model, \mosaic{} finds solutions that are physically grounded.
 
Our contributions are:
\begin{itemize}
    \item A feasibility-driven planning framework, \mosaic{}, that shifts from goal-directed to a multi-directional search anchored in regions of high skill competence, along with its theoretical foundations.
    \item A physics-informed reasoning approach that uses high-fidelity simulation to ground planning decisions in physical reality through in-the-loop evaluation.
    \item Extensive validation in both simulation and the real world, demonstrating that \mosaic{} outperforms existing methods in performance, generalization, and scalability on complex manipulation tasks.
\end{itemize}

%% file: tex/related_work.tex
\section{Related Work}
We categorize prior works into four main areas: Model-Based Planning with Simulation, Task and Motion Planning, Single Policy Methods, and Skill-Based Methods.

\textbf{Model-Based Planning with Simulation} remains fundamental for autonomous decision-making in complex environments, enabling systematic reasoning about action consequences and exploration of alternative strategies before execution. 
This is particularly crucial for long-horizon manipulation tasks where myopic decisions can lead to dead ends or suboptimal solutions. 
To plan effectively, we need a world model that can estimate the outcomes of different actions. 
Physics simulators have seen remarkable advances, with modern platforms like Isaac Sim \cite{NVIDIA_Isaac_Sim}, Genesis \cite{Genesis}, Sapien \cite{Xiang_2020_SAPIEN}, and MuJoCo \cite{MuJoCo} now capable of modeling complex contact dynamics with unprecedented accuracy and efficiency. 
These advances have enabled zero-shot sim2real transfer \cite{zakka2025mujoco} and real2sim2real \cite{torne2024reconciling} workflows that allow robots to perform complex tasks in the real world.
A growing body of work has demonstrated the effectiveness of integrating physics simulators directly into the planning loop. 
\citet{saleem2020planning, saxena2021manipulation, saxena2023planning} developed algorithms for manipulation planning among movable objects using selective physics-based simulation evaluations, solving complex manipulation problems involving both prehensile and non-prehensile manipulation. 
There is also significant work on using physics simulators and differentiable physics simulators for optimization through contact-rich scenarios \cite{natarajan2023longhorizon, le2024fast, le2025model, shen2025differentiablegpuparallelizedtaskmotion}.
While learned world models \cite{li2025robotic, assran2025v} show promise for efficient world modeling, physics simulators currently provide the most reliable predictions for manipulation tasks involving complex physical interactions. 
In this work, \mosaic{} leverages physics simulation as a high-fidelity world model to ground planning decisions in physical reality. \mosaic{} is designed to accommodate any world model that can evaluate skill outcomes, and will continue improving as world models increase in fidelity and efficiency.

\textbf{Task and Motion Planning (TAMP)} addresses long-horizon planning problems by interleaving abstract symbolic task planning with geometric motion planning. 
TAMP systems require users to define a symbolic task specification and a symbolic world state that can be augmented by actions \cite{garrett2021integrated}. 
These systems are powerful but require manual engineering and privileged task-specific information for determining the symbolic representations and available transitions \cite{crosby2016skiros}. 
While there is a significant body of work that aims to learn different TAMP modules to avoid manual engineering \cite{konidaris2018skills, mao2022pdsketch, kim2018guiding, wang2021learning, migimatsu2022symbolic, silver2022learning, liang2022search,fang2024dimsam}, designing appropriate learning modules that interface with existing symbolic planners and collecting training data to learn accurate symbolic abstractions is challenging. 
Recent hybrid approaches, such as \cite{zhou2024spire}, combine TAMP with learning by decomposing tasks symbolically and training policies for subtasks. 
\mosaic{} takes a different approach and eliminates the need for symbolic task specifications.
Instead, it relies on physics simulators to directly forward simulate any skills considered during its search -- an attractive alternative made possible by the efficiency and high-fidelity of modern simulators.

\textbf{Single Policy Methods} learn to solve tasks under specified goal conditions. Diffusion models have been used for this purpose \cite{janner2022diffuser, chi2024diffusionpolicy}, showing strong performance in manipulation but struggling with generalization and long-horizon reasoning. 
In reinforcement learning, automatic goal generation \cite{florensa2018automatic, forestier2022intrinsically} and option discovery \cite{veeriah2021discovery} incorporate domain knowledge via structural priors into the learning process to guide exploration and improve sample efficiency. 
These methods make full-horizon learning more tractable by injecting task-specific information. 
However, policy learning methods typically succeed only in short-horizon tasks and struggle with the long-horizon reasoning and skill composition required for complex manipulation due to large data requirements and poor generalization without substantial retraining. 
In contrast, \mosaic{} leverages simple, task-agnostic skill primitives to solve diverse long-horizon tasks through intelligent composition rather than end-to-end learning.

\textbf{Skills-based Methods} aim to compose skills (e.g., learned policies, analytic controllers) to solve long-horizon tasks. 
They typically address two challenges: selecting \textit{which skills} to compose (a discrete problem) and \textit{how to parameterize} them (often a continuous problem). 
A central approach is skill chaining \cite{konidaris2009skillchaining}, which, building on the options framework \cite{sutton1999options}, incrementally discovers local policies that achieve subgoals \cite{bagaria2019option, bagaria2021robustly, bagaria2021skill, xu2021daf}. 
Other methods \cite{nasiriany2022augmenting} explicitly select skills and parameters, or learn skills as diffusion models~\cite{mishra2023generative, mishra2024generative}. 
However, the sequential structure of these frameworks forces skills to define initiation, termination, and effect sets, or assumes access to plan skeletons, limiting their flexibility and reuse across tasks. 
\citet{sivaramakrishnan2024roadmaps} relax some of these requirements by learning a goal-conditioned dynamics policy in an obstacle-free space, then building a roadmap to guide planning with the learned controller. Similarly, \cite{lu2020reset} use sampling-based planners to explore the abstract space of higher-order skills in lifelong RL.

A fundamental limitation shared by existing skill-based methods--and in most planning approaches--is their reliance on \textit{goal-directed search}: forward methods choose skills based on reachable termination sets from the start state, while backward methods select skills based on initialization sets leading to the goal state. 
Both schemes constrain exploration to local neighborhoods around current or goal states, making it difficult to identify distant but executable skills that could aid task completion. 
This approach can be particularly inefficient when solutions depend on non-obvious intermediate steps that offer no immediate indication of progress toward the goal.
Critically, skills are \textit{imperfect}--they succeed under certain conditions but may fail in others due to environmental constraints, object configurations, or inherent limitations in their training data. 
To the best of our knowledge, no existing planning framework explicitly reasons about where skills are likely to succeed and directs planning efforts accordingly.

\mosaic{} differs by addressing this gap through a feasibility-driven, multi-directional search that plans with imperfect skills. 
Rather than exploring skill sequences in a goal-directed fashion, \mosaic{} solves complex long-horizon tasks by explicitly reasoning about skill competence and discovering regions where skills are most likely to succeed.

%% file: tex/problem_formulation.tex
\section{Problem Definition: Skill-Centric Planning}
\label{sec:problem_definition}

We seek to find a sequence of parameterized skills, out of a given library of available skills, whose execution would modify a world state from its current specification to a specification satisfying a goal condition. 

Formally, let $\mathcal{Q_R} \subseteq \mathbb{R}^n$ denote the configuration space of a robot $\mathcal{R}$ with $n$ degrees-of-freedom (DOF), and let $\mathcal{X} \subseteq \mathbb{R}^m$ be the state space of the planning problem, often called the \textit{world state}, where $n \leq m$. To facilitate set membership checks for states (e.g., whether a state is a goal state or belongs to some equivalence class), let us define \textit{binary conditions} $\xi: \mathcal{X} \rightarrow {0, 1}$ with $\xi \in \Xi$.
We define a \textit{trajectory} $\ve{\tau}$ as a mapping $\ve{\tau}: [0,1] \rightarrow \mathcal{X}$, and $\Tau$ as the trajectory space. The space of \textit{parameterized skills}, i.e., motor controllers that map skill parameters $\theta$ to trajectories $\ve{\tau} \in \Tau$~\cite{ames2018learning}, is termed the skill space $\mathcal{A}$.

Our definition of \textit{parameterized skills} is related to the \textit{options} framework \cite{sutton1999options}. A skill $\sigma$ is a temporally extended action, represented as a tuple $(\pi_\sigma, \Theta_\sigma)$, where $\pi_\sigma$ is the skill policy that returns the probability of taking action $a$ in state $x$ given parameters $\theta \in \Theta_\sigma$ via $\pi_\sigma(a | x, \theta)$. Actions $a$ may denote low-level control commands (e.g., joint velocities) or higher-level motion segments (e.g., waypoint sequences). The outcome of a skill is a trajectory $\ve{\tau} \in \Tau$.

Crucially, this does not imply that skills must always be executed in an open-loop fashion. 
Trajectories serve as the planning representation of skills, but execution can remain closed-loop. Skills may be realized as reactive policies (e.g., diffusion policies~\cite{chi2024diffusionpolicy}), in which case their trajectories are obtained by rollout; as deterministic planners, where trajectories are generated procedurally; or as hybrid methods that combine both. Thus, skills provide a unified interface for planning—via trajectories—while retaining flexibility in execution. We next distinguish between two types of skills tailored to our framework.

\begin{mydef}
    \label{def:generator}
    \textbf{\textit{Generators}}, denoted as $\mathcal{G}_i : \Theta_{\mathcal{G}_i} \rightarrow \Tau$, are parameterized skills 
    that generate trajectories without requiring specified start and goal states.
\end{mydef}
\begin{mydef}
    \label{def:connector}
    \textbf{\textit{Connectors}}, denoted as $\mathcal{C}_j : \Xi \times \Xi \times \Theta_{\mathcal{C}_j} \rightarrow \Tau$, are \textit{conditional} parameterized skills that generate trajectories conditioned on specified start and goal conditions.
    \footnote{A common start and goal condition for connector skills is an equality condition to given states $x'$ and $x''$. Therefore, for brevity when the context is clear, we interchangeably write
    $\mathcal{C}_j : \mathcal{X} \times \mathcal{X} \times \Theta_{\mathcal{C}_j} \rightarrow \Tau$
    and imply the conditions $\mathds{1}_{\{x'\}}$ and $\mathds{1}_{\{x''\}}$.}
\end{mydef}
With these definitions, we formally define the skill space as $\mathcal{A} = \bigcup_i \mathcal{G}_i \cup \bigcup_j \mathcal{C}_j$.
Given a start state $x_{\text{start}}$ and goal condition function $ \xi_\text{goal}:\mathcal{X} \rightarrow \{0, 1\}$ 
the objective of the planning problem is to find a sequence of $N$ skills $\{\sigma_i \mid \sigma_i \in \mathcal{A}\}_{i=1}^N$ and associated parameters $\{\theta_i \mid \theta_i \in \Theta_{\sigma_i}\}_{i=1}^N$ that produce the sequence of trajectories $\Pi = \{\ve{\tau}_1, \ve{\tau}_2,\dots, \ve{\tau}_N\}$ that satisfies the following conditions:
\begin{equation}
    \label{eq:start_condition}
    \ve{\tau}_1(0) = x_{\text{start}}
\end{equation}
\begin{equation}
    \label{eq:goal_condition}
    \xi_\text{goal}(\ve{\tau}_N(1)) = 1
\end{equation}
\begin{equation}
    \label{eq:continuity_condition}
    \ve{\tau}_i(1) = \ve{\tau}_{i+1}(0) \quad \forall i \in \{1, \dots, N-1\}
\end{equation}

That is, the solution sequence must satisfy three key properties: the initial state of the first trajectory must be equal to the start state (Eq. \ref{eq:start_condition}), adjacent trajectories must connect continuously 
(Eq. \ref{eq:continuity_condition}), and the final state of the last trajectory must satisfy the goal condition (Eq. \ref{eq:goal_condition}). Skills produce only \textit{valid} trajectories $\ve{\tau}_i$: not causing collisions between the robot and static obstacles and respecting robot dynamics.
Additional constraints (e.g., motion smoothness) and optimization objectives (e.g., execution time) may be imposed on the solution.

\subsection{Skill-centric Manipulation Planning}

Robotic manipulation refers to the process of altering the state of objects through physical interaction. A fundamental characteristic of manipulation planning is that the system is typically underactuated -- the number of actuated DOFs is smaller than the dimension of the state space $\mathcal{X}$. 
For each movable object $\mathcal{O}_i$ we define a configuration space $Q_{\mathcal{O}_i} \subseteq SE(3)$. Given 
a robot's configuration space $\mathcal{Q_R}$, 
the complete state space is then defined as the product of all configuration spaces, $\mathcal{X} = \mathcal{Q_R} \times Q_{\mathcal{O}_1} \times \dots \times Q_{\mathcal{O}_k}$, where $k$ is the number of movable objects. 
Since movable objects are included in the state space and are not directly actuated, the state space dimension $m$ is generally greater than the robot's configuration space dimension $n$,
resulting in an underactuated system. 
Therefore, generating a new state during planning requires specifying not only the robot's configuration but also the configurations of all movable objects. 
To accurately capture how robot actions affect the environment, we rely on a model--in our case, a physics simulator--that can predict the outcomes of these interactions.

%% file: tex/mosaic.tex
\section{\mosaic{}}
\label{sec:mosaic}

The core concept of \mosaic{} is the construction of a directed multigraph, which we call a \textit{mosaic graph}. 
In this graph, nodes are tuples of skills, parameters and the estimated generated trajectories, and edges are tuples of skills, parameters, and boundary conditions. 
Nodes are created with \textit{generator} skills, which produce local behavior trajectories, and edges are created with \textit{connector} skills, which link the generated local trajectories.
Since constructing this graph naively by alternating between generator and connector skills can result in excessive computational overhead (Sec.~\ref{section:experimental-results}), 
the \mosaic{} algorithm employs a guidance module, called the \textit{oracle}, which orchestrates the construction process by selecting appropriate skills and determining which parts of the mosaic graph to connect. 
In this section, we first outline the algorithm and then provide an overview of the oracle.

\subsection{Algorithmic Approach}
\label{sec:alg_app}

\begin{algorithm}[t]
    \SetCustomAlgoRuledWidth{0.58\textwidth}  
        \SetKwFunction{DirectedMultigraph}{DirectedMultigraph()}
        \SetKwFunction{NODES}{NODES}
        \SetKwFunction{EDGES}{EDGES}
        \SetKwFunction{AddEdge}{AddEdge}
        \SetKwFunction{AddNode}{AddNode}
        \SetKwFunction{SampleParameters}{SampleParameters}
        \SetKwFunction{SampleGoalState}{SampleGoalState}
        \SetKwFunction{goalNodes}{goalNodes}
        \SetKwFunction{ChooseSkill}{ChooseSkill}
        \SetKwFunction{HasPath}{HasPath}
        \SetKwFunction{ChooseCondsToConnect}{ChooseCondsToConnect}
        \SetKwFunction{Cost}{Cost}
        \SetKwFunction{ShortestPath}{ShortestPath}
        
        \scriptsize
        \caption{\mosaic{}}
        \label{alg:mosaic}
        \KwIn{Start state $x_{\text{start}} \in \mathcal{X}$ \newline 
                Goal termination condition function $\xi_{goal}: \mathcal{X} \rightarrow \{0,1\}$ \newline 
                Skill library $\Sigma = \{\sigma\}_{m=1}^M$ \newline
                Oracle $O$ for skill and trajectory selection 
                }
        \KwOut{Sequence of skills and their parameters $\Pi$} 
        \vspace{4pt}
        
        $\mathcal{M}$ = \DirectedMultigraph 
        \label{line:mosaic}
        
        $\mathcal{G} \leftarrow \{ \mathcal{G}_i \in \Sigma$\} \tcp*[f]{\color{orange} \scriptsize Generators container} 
        
        $\mathcal{C} \leftarrow \{ \mathcal{C}_j \in \Sigma$\} \tcp*[f]{\color{orange} \scriptsize Connectors container}
        \vspace{5pt}
        
        \While{$\mathcal{M}$.\NODES $ = \emptyset$}{\label{line:init-generators}
            \For{$\sigma \in \mathcal{G}$}{
                $\theta \gets O.\SampleParameters (\sigma)$

                $\ve{\tau} \leftarrow \sigma(\theta)$ \tcp{\color{orange} \scriptsize Estimatd, valid trajectories}
                
                \If{$\ve{\tau} \neq \emptyset$}{
                    $\mathcal{M}.\AddNode((\sigma, \theta, \ve{\tau}),\Cost(\ve{\tau}))$
                }
            }
        }\label{line:end-generators}
    
        $\ve{\tau}_{\text{start}} \gets \{ x_{\text{start}} \}$
        
        $\mathcal{M}.\AddNode((\emptyset, \emptyset, \ve{\tau}_{\text{start}}), 0)$  \tcp*{\color{orange} \scriptsize Add start state as a node, with zero cost}
    
        \vspace{4pt}
    
        \While{$\neg \mathcal{M}.\HasPath(x_{\text{start}}, \{x \mid \xi_{goal}(x) = 1\})$}{\label{line:mainloop}
            \tcp{\color{orange} \scriptsize Oracle selects next skill to apply}
            $\sigma \gets O.\ChooseSkill(\Sigma, \mathcal{M})$ \label{line:choose-skill} 
            
            $\theta \gets O.\SampleParameters(\sigma)$
            
            \eIf{$\sigma \in \mathcal{C}$}{\label{line:connectorloop}
                \tcp{\color{orange} \scriptsize Apply connector skill}
                $\xi_0, \xi_1 \gets O.\ChooseCondsToConnect(\mathcal{M})$ \label{line:chooseconds}
                
                $\ve{\tau} \gets \sigma(\xi_0, \xi_1, \theta)$  \label{line:get_trajs1}
                
                \If{$\ve{\tau} \neq \emptyset$}{
                    $\mathcal{M}.\AddEdge((\sigma, \xi_0, \xi_1, \theta), \Cost(\ve{\tau}))$
                }
            }{\label{line:connectorendloop}
                \tcp{\color{orange} \scriptsize Apply generator skill}
                $\ve{\tau}\gets \sigma(\theta)$ \label{line:generatorloop} \label{line:get_trajs2}
                
                \If{$\ve{\tau} \neq \emptyset$}{
                    $\mathcal{M}.\AddNode((\sigma, \theta, \ve{\tau}), \Cost(\ve{\tau}))$
                }
            }\label{line:generatorloopend}
        }\label{line:mainloopend}
        
        \tcp{\color{orange} \scriptsize Least-cost path to any goal state}
        \KwRet $\mathcal{M}.\ShortestPath(x_{\text{start}}, \{x \mid \xi_{goal}(x) = 1\})$ \label{line:shortest-path}
\end{algorithm}

The \mosaic{} algorithm (Alg. \ref{alg:mosaic}) finds a sequence of skills whose application to the given start state $x_\text{start}$ results in a state satisfying the goal condition $\xi_\text{goal}$. \mosaic{} does so by exploring the search space under the guidance of an oracle $O$, using a skill library $\Sigma$ composed of skills $\sigma$ that act as connectors ($\sigma \in \mathcal{C}$), generators ($\sigma \in \mathcal{G}$), or both.

The algorithm starts by initializing the graph (line \ref{line:mosaic}) and invoking all available generators -- adding all valid skills and their associated parameters as disconnected nodes in the mosaic graph (lines \ref{line:init-generators}–\ref{line:end-generators}). If no nodes were added to the graph due to generator failures, the algorithm repeatedly queries the generators with different parameters until it finds at least one valid node.

The main loop of \mosaic{} (lines \ref{line:mainloop}–\ref{line:mainloopend}) begins with the oracle selecting a skill to invoke next (line \ref{line:choose-skill}), which can be either a generator or a connector, and samples its parameters. If the selected skill is a connector, the oracle assigns it a start condition and a goal condition (line \ref{line:chooseconds}). Most often, the connector skill chooses two skill trajectories $\ve{\tau}_0$ and $\ve{\tau}_1$ from the mosaic and sets  $\xi_0 := \mathds{1}_{\{\ve{\tau}_0(1)\}}$ and $\xi_1 := \mathds{1}_{\{\ve{\tau}_1(0)\}}$ to attempt connecting them. Otherwise, the connector will attempt to connect a chosen node to the goal condition by setting $\xi_1 := \xi_\text{goal}$.
Next, the selected skill $\sigma$ is invoked, and only the skills and their parameters that produce valid trajectories are added to the mosaic graph -- as edges in the case of a connector (line \ref{line:get_trajs1}) or as nodes in the case of a generator (line \ref{line:get_trajs2}). If a skill may lead to more than one trajectory due to stochasticity of its corresponding policy, multiple trajectories are generated, e.g., via neural network batch inference or parallel computation on CPU cores, and \mosaic{} leverages this information to estimate a confidence value (reflected in the cost value) associated with the selected skill by considering the fraction of invalid trajectories in the batch
(lines \ref{line:init-generators}–\ref{line:end-generators}, \ref{line:connectorloop}–\ref{line:connectorendloop} and \ref{line:generatorloop}–\ref{line:generatorloopend}).
Finally, if the mosaic graph contains a path between the start state and a state that satisfies the goal condition, the algorithm returns (line \ref{line:shortest-path}).

\begin{figure*}[tb]
    \centering
    \subcaptionbox{Transport.\label{fig:exp_no_obs_sim}}[0.32\textwidth]{%
        \includegraphics[width=0.49\linewidth]{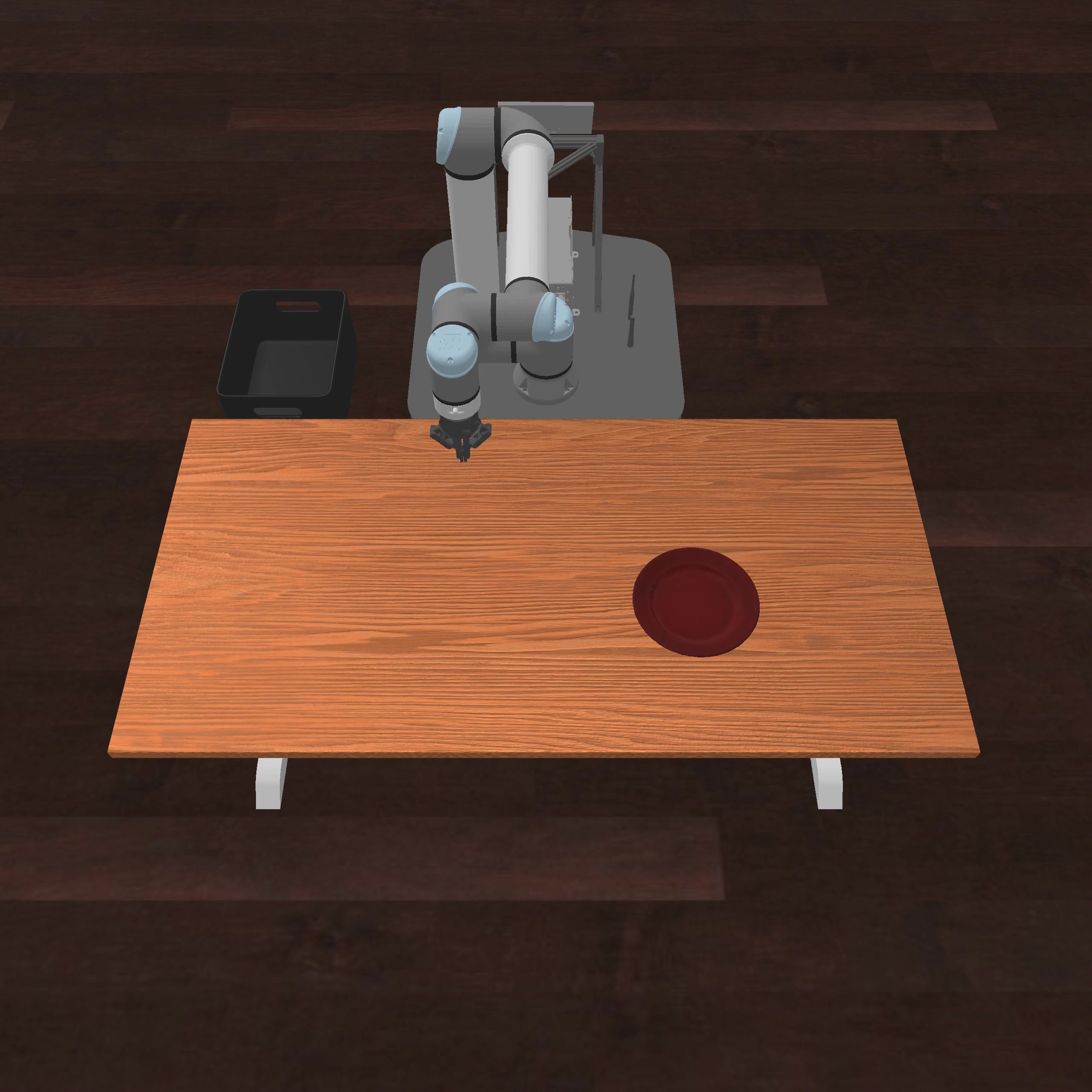} \includegraphics[width=0.49\linewidth]{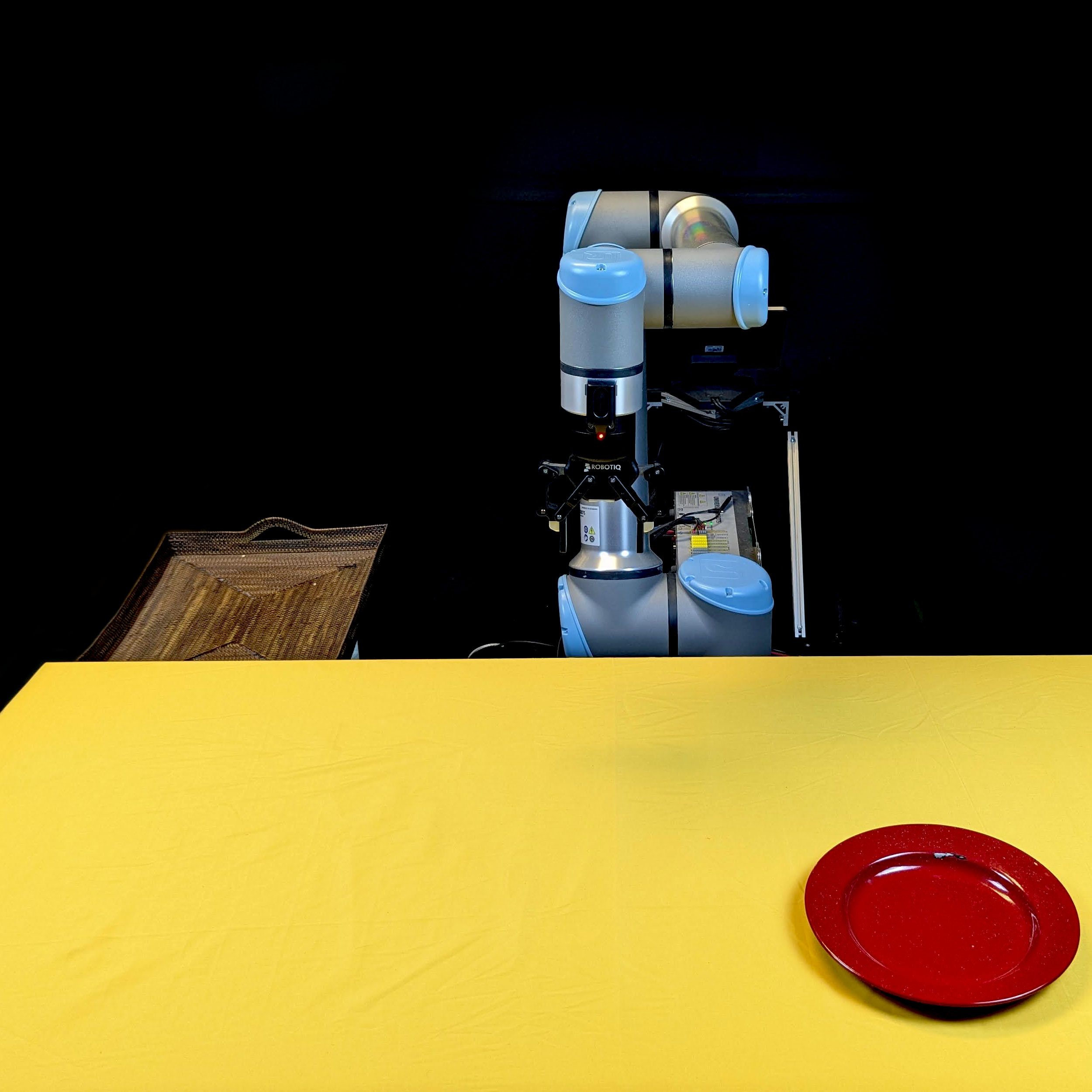}
    }
    \hfill   
    \subcaptionbox{Transport in Clutter.}[0.32\textwidth]{%
        \includegraphics[width=0.49\linewidth]{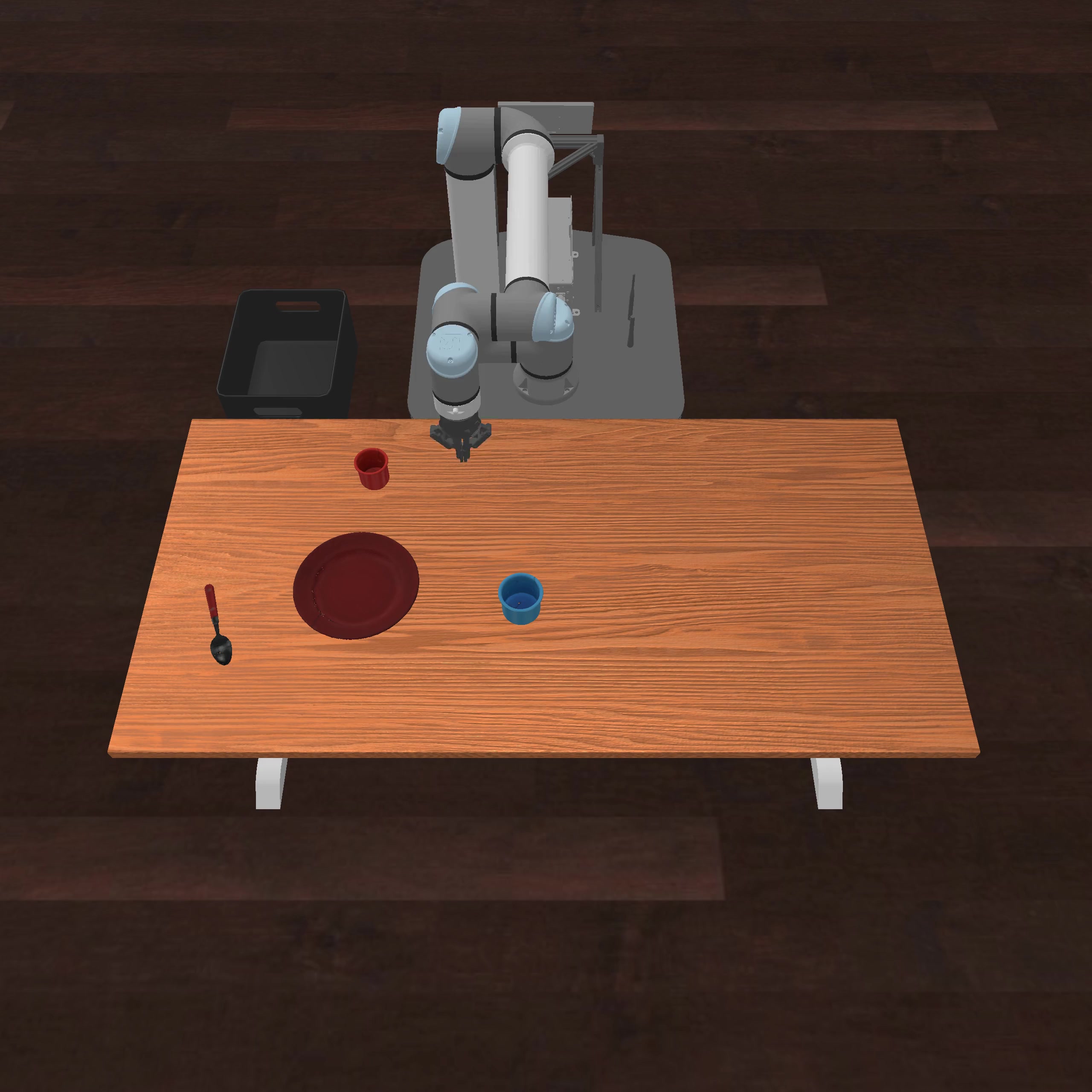}
        \includegraphics[width=0.49\linewidth]{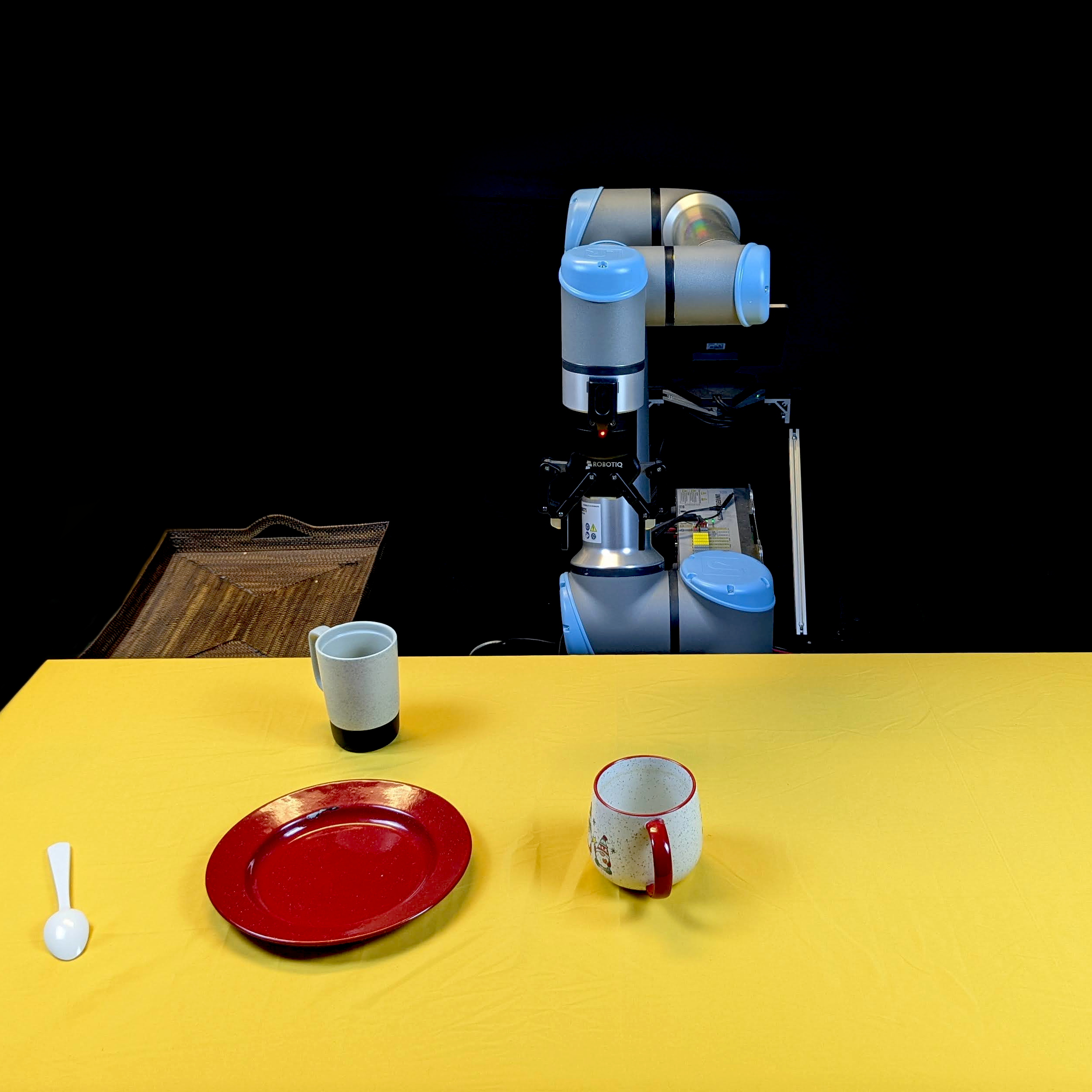}
    }
    \hfill   
    \subcaptionbox{Transport Among Movable Objects.}[0.32\textwidth]{%
        \includegraphics[width=0.49\linewidth]{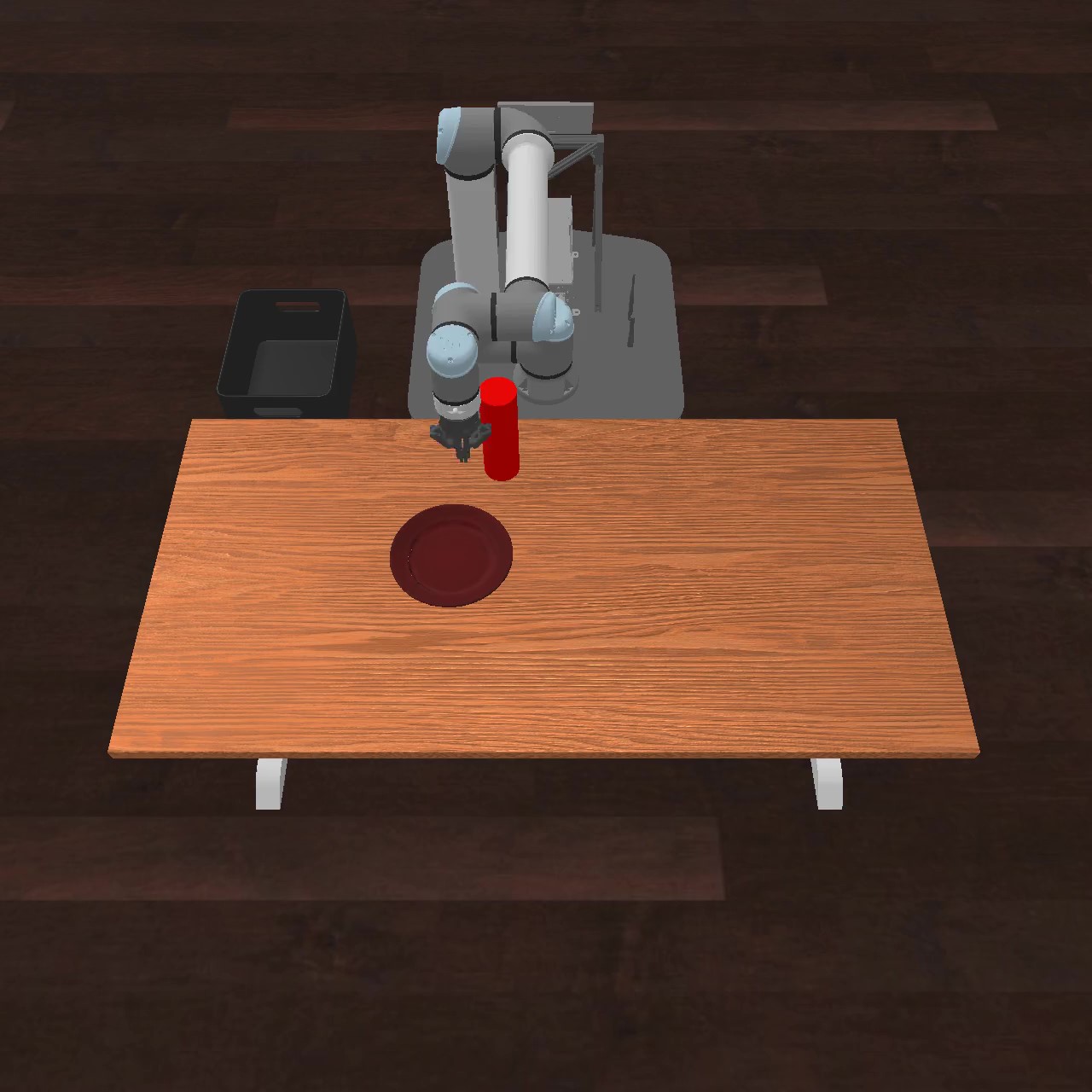}
        \includegraphics[width=0.49\linewidth]{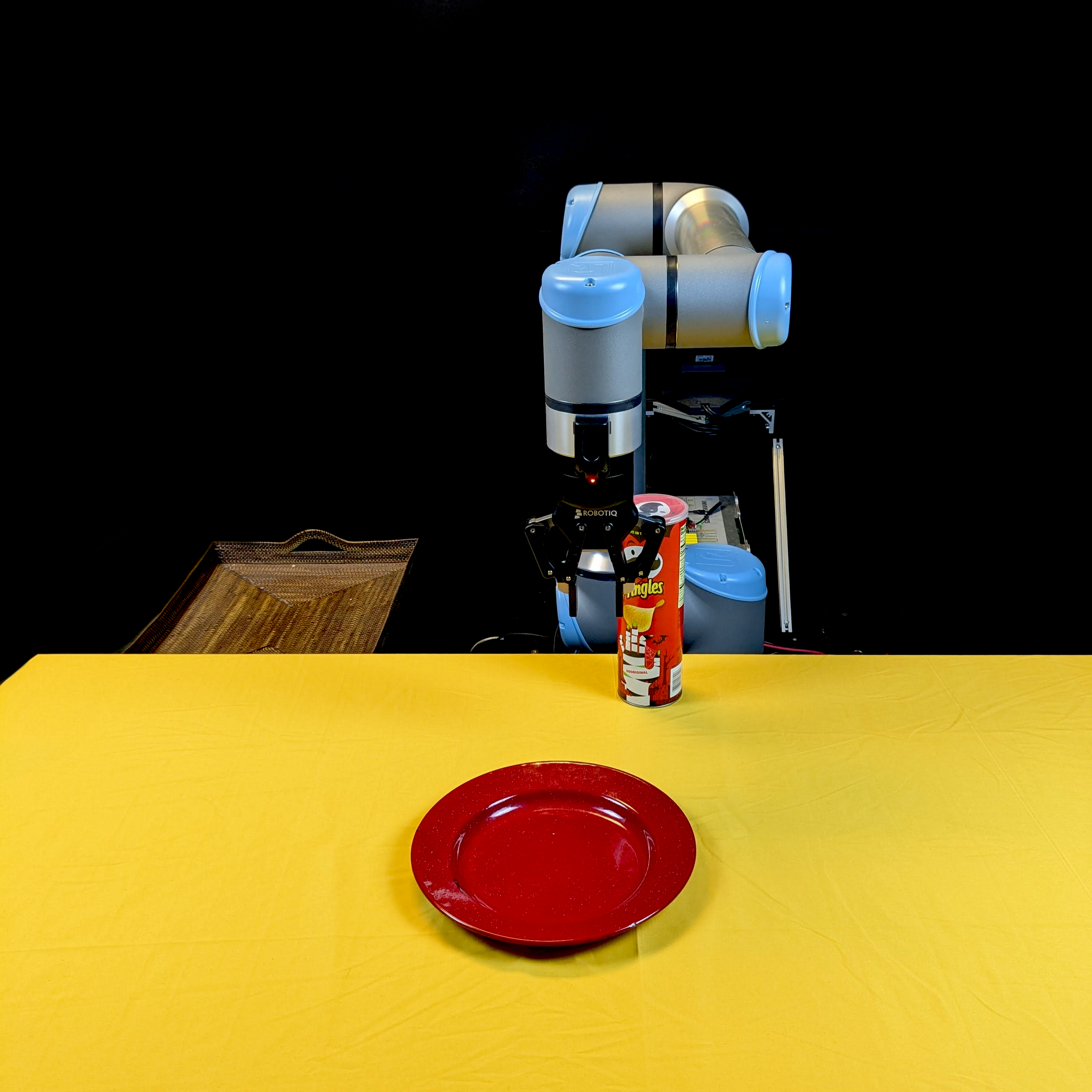}
    }
    \caption{
        Simulation setups and their real-world counterparts. Across all scenarios, the robot must place the plate into the bin. In Scenario 1: Transport (a), the robot must push the plate to the edge in order to pick it up. Scenario 2: Transport in Clutter (b) includes additional objects on the table. The robot must move the plate to the edge without displacing other objects. Scenario 3: Transport Among Movable Objects (c) allows the robot to interact with more objects on the table. The robot discovered the need to clear space for manipulating the plate by moving the chips can elsewhere.
    }
    \label{fig:experimental-setup}
\end{figure*}

\subsection{Oracle}

\mosaic{} explores the vast space of parameterized skills (i.e., expands its implicit multigraph) using its guidance module, which we term the \emph{oracle}.
The oracle's role is to orchestrate the search by balancing exploration--using \textit{Generators} to discover new ``islands of competence''--and exploitation--using \textit{Connectors} to build edges between existing islands and find a solution. 
By making principled decisions at each step, the oracle directs the search toward robust, task-relevant skills.

The oracle within \mosaic{} can be implemented in a variety of ways, and may be specialized with task-specific knowledge or leverage advancements in other fields (e.g., the oracle can be a large language model). Aiming to remain independent from specific tasks in this work, we propose an effective domain-independent statistical oracle that makes decisions based on the evolving structure of the search graph.

Our proposed oracle module is general and can be used with any library of skills. Specifically, when invoked, our oracle first decides whether to choose a Generator or a Connector based on the graph's connectivity; the probability of selecting a Connector increases as the ratio of nodes $N$ to edges $E$ in the graph grows. 
Once a skill type is chosen, a specific skill $\sigma$ is selected by maximizing a score $U(\sigma_i)$ that balances its past performance with an exploration bonus:

\[U(\sigma_i) = \alpha s_{\sigma_i} + (1-\alpha) \sqrt{\ln\frac{\sum_j (t_{\sigma_j} + 1)}{t_{\sigma_i} + 1}} + n \]

\noindent with the $s_{\sigma}$ being the success rate of skill $\sigma$, $t_{\sigma}$ its invocation count, $\alpha$ a weighting parameter, and $n \sim \mathcal{N}(0, 1)$ a noise term for stochasticity.

When $\sigma$ is a generator, it unconditionally creates a new node in the \mosaic{} graph, and when it is a Connector, the oracle provides it with boundary conditions to connect.
In the latter, the oracle may randomly select a node to connect locally to one of its neighbors\footnote{We define a distance between nodes $x$ and $x'$ as the pose difference of robots and objects in the final state in $x$ and initial state in $x'$.} -- promoting unbiased exploration, or exploit the structure of the graph by selecting
nodes already reachable from the start state or the goal. 
To avoid redundant effort, the oracle penalizes pairs with repeated failed connection attempts by artificially increasing their distance, discouraging their selection in future nearest-neighbor searches. 
This design provides an effective, statistically-grounded guidance strategy that requires no task-specific tuning.

\subsection{Theoretical Analysis: Probabilistic Completeness}
\label{sec:theory}

In this section, we formally define probabilistic completeness (PC) in the context of skill-centric planning and provide a proof sketch demonstrating why \mosaic{} is PC. For the purposes of these proofs, we assume deterministic skill-trajectory generation, i.e., each skill and parameter pair $\langle \sigma, \theta \rangle$ maps to a consistent trajectory (or a batch of trajectories). To define PC with respect to a given skill library, we first introduce the concept of a \textit{feasible} solution:

\begin{mydef}
    \label{def:feasibility}
    Given a start state $x_{start}$, a goal condition function $\xi_\text{goal}: \mathcal{X} \rightarrow \{0, 1\}$, and a library of skills $\Sigma = \{ \sigma_m \in  \mathcal{A}\}_{m=1}^M$, a solution trajectory is said to be \textbf{feasible} under $\Sigma$ if it is decomposable into a sequence of trajectory segments, $\Pi = \{\ve{\tau}_1, \dots, \ve{\tau}_N\}$ such that $\forall \ve{\tau}_i \in \Pi, \exists \sigma_i \in \Sigma \text{ with parameters } \theta \in \Theta_{\sigma_i}$ that can generate $\ve{\tau}_i$. 
\end{mydef}

Given the definition of feasibility, a PC algorithm must guarantee that, asymptotically, it will discover a sequence of trajectories that compose a feasible solution if one exists:

\begin{mydef}
    \label{def:Skill-centric Probabilistic Completeness}
    Let $\Pi_k^{\text{\textbf{ALG}}}$ be the set of trajectories discovered by a skill-centric planning algorithm \text{\textbf{ALG}} at its $k$-th iteration. Additionally, let $\Pi^*$ be the set of all feasible solutions for the planning problem.  We call an algorithm \textbf{probabilistically complete} if $\lim\limits_{k \rightarrow \infty } P(\exists \Pi \in \Pi_k^{\text{ALG}} \quad | \quad \Pi \in \Pi^*) = 1$.
\end{mydef}

To establish PC, \mosaic{} must satisfy two conditions. As the number of iterations $k \rightarrow \infty$, it must (1) invoke all generator skills ($\mathcal{G} \in \Sigma$) with every parameter configuration to explore all potential trajectory segments, and (2) attempt to connect every disconnected node using all connector skills ($\mathcal{C} \in \Sigma$) with all parameters. 
This is ensured by the fact that the oracle assigns non-zero probabilities to selecting any skill and sampling parameters randomly. The oracle's sampling strategy guarantees that, given infinite time, all possible skill combinations and parameter configurations will be explored. Thus, all generators and connectors are eventually used across all configurations, guaranteeing that \mosaic{} explores all trajectories and connections, and finds a solution if one exists.

%% file: tex/experimental_analysis.tex
\section{Experimental Analysis}
\label{sec:experimental_analysis}

\begin{figure*}[h!]
    \centering
    \includegraphics[width=0.99\textwidth]{images/planner_comparison_no_obs_no_xaxs.png}
    \includegraphics[width=0.99\textwidth]{images/planner_comparison_w_obs_no_xaxs.png}
    \includegraphics[width=0.99\textwidth]{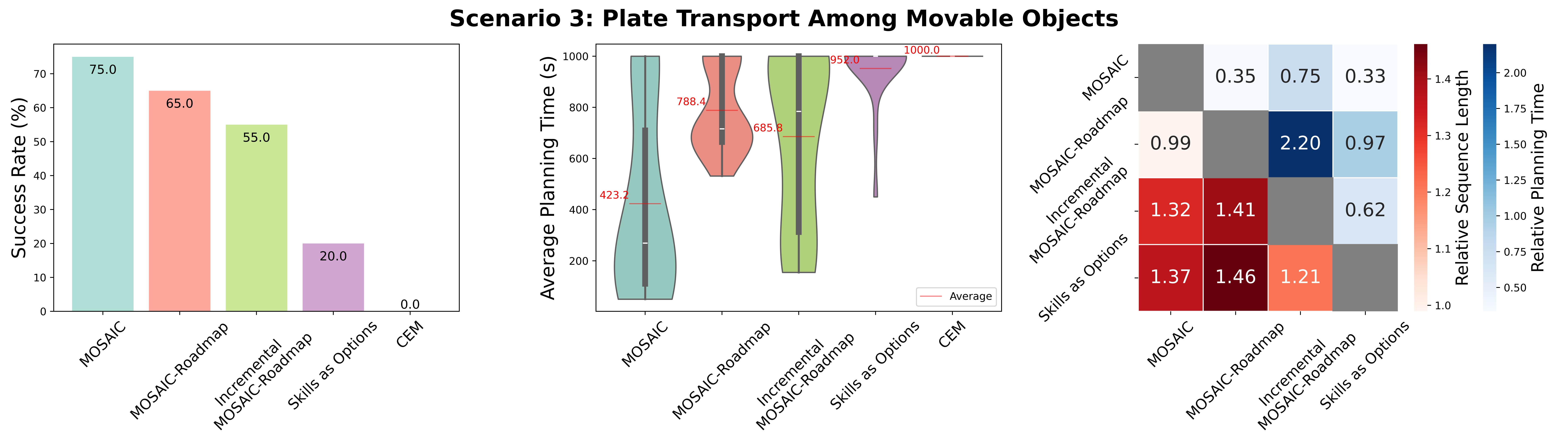}
    \caption{
    Algorithms comparison across experimental scenarios.
    \textbf{Left}: Success rates. \textbf{Middle}: Planning time density with median, IQR, and average. \textbf{Right}: Head-to-head comparison on tests both algorithms solved. Upper-right shows relative planning times; lower-left shows relative sequence lengths. Each cell compares the ``row algorithm" to the ``column algorithm." For \mosaic{}, lower values are better in the first row, and higher values are better in the first column.
    }
    \label{fig:planner_comparison}
\end{figure*}

We evaluate \mosaic{} in diverse simulated environments and validate on real UR10e robot, using RealSense for RGB-D pose estimation. We test three manipulation scenarios of increasing complexity, providing only the goal condition (``plate in bin") and  a set of generic skills. Figure~\ref{fig:experimental-setup} depicts the setups; the accompanying video shows the hardware executions.

\emph{Scenario 1: Basic Transport} places a plate deep on the table where it cannot be grasped directly due to its geometry. The planner must discover a sequence that first reorients the plate (e.g., by sliding it outward), then grasps it from the side, and places it in the bin.
\emph{Scenario 2: Transport in Clutter} extends Scenario 1 by adding static obstacles around the plate. The planner must discover a skill sequence that navigates through clutter without collisions.
\emph{Scenario 3: Transport Among Movable Objects} introduces a second movable object that may obstruct plate access. The planner can choose to work around this object or manipulate it, expanding the search space to include two-object interactions.

\subsection{Skill Library}
We tested \mosaic{} with a library of four skills that can serve as generators ($\mathcal{G}$), connectors ($\mathcal{C}$), or both:
\textbf{Push} ($\mathcal{G}+\mathcal{C}$) uses a learned diffusion policy \cite{chi2024diffusionpolicy} to generate pushing motions up to 25cm with 70\% success rate. As a generator, it creates new world states with local push trajectories; as a connector, it moves objects between poses across world states. This skill uses a motion planner as a subroutine to arrive at pre-push configurations.
\textbf{Pick} ($\mathcal{G}$) computes grasp poses based on object geometry and arm kinematics, then executes screw-based approach-grasp-retract trajectories.
\textbf{Transport} ($\mathcal{C}$) moves grasped objects to satisfy goal conditions using motion planning \cite{RRT-Connect, OMPL2012}. The skill fails if the object is not grasped or the goal isn't met.
\textbf{Rearrange} ($\mathcal{C}$, Scenario 3 only) combines pick-and-place and push policies to reposition multiple objects by determining the appropriate policy for each object.
During planning, the skills are rolled-out in the Sapien physics engine \citep{Xiang_2020_SAPIEN,gu2023maniskill2} to verify contact outcomes and prune infeasible plans. 
Additional implementation details appear in the Appendix (Sec.\ref{sec:appx:skills}).

\subsection{Evaluation Metrics}
We measure algorithm performance primarily by \emph{success rate}, defined as the fraction of trials where the planner reaches a goal-satisfying state $\xi_{\text{goal}}$. Additionally, we record the \emph{plan length}, measured as the number of skills comprising the solution, serving as a proxy for task complexity and solution quality. Lastly, we measure each algorithm’s \emph{planning time} to evaluate its efficiency.

\subsection{Baseline Algorithms}

We compare \mosaic{} to four long-horizon planners that compose skill primitives:
\textbf{Skills as Options} implements sequential skill chaining \cite{konidaris2009skillchaining}, exploring skills incrementally with each new skill starting where the previous ended. It performs breadth-first search over world states by creating an action space of start-conditioned generator skills.
\textbf{CEM} \cite{cem} uses receding horizon planning, sampling skill-parameter sequences, selecting top candidates, and refining the sampling distribution. It executes the first skill of the best sequence, updates the world state, and replans until success or timeout.
\textbf{\mosaic{}-Roadmap} removes sequential dependencies by generating skill trajectories freely and connecting them via a PRM-inspired roadmap \cite{PRM}. The algorithm first builds a roadmap using both generator and connector skills, then attempts to connect the start and goal nodes. If successful, it uses Dijkstra's algorithm \cite{dijkstra1959note} to find a path; otherwise, it reports failure. 
\textbf{Incremental \mosaic{}-Roadmap} extends \mosaic{}-Roadmap by iteratively adding nodes and edges to the roadmap until start and goal nodes are connected, ensuring continuous progress toward a solution even if the initial roadmap doesn't yeild a path.

\subsection{Experimental Results}
\label{section:experimental-results}

Our experiments address three key questions: (1) Can \mosaic{} effectively solve long-horizon skill-centric planning problems? (2) How important is the oracle module in \mosaic{}'s performance? (3) How does \mosaic{}'s skill discovery and composition strategy compare to traditional directed approaches? The results, shown in Fig. \ref{fig:planner_comparison}, demonstrate that \mosaic{} consistently achieves high success rates across all scenarios while maintaining competitive solution quality and planning times. The performance of the baselines reveals important insights about different planning approaches: \textit{Skills as Options}, representing current sequential methods, performs well on simple tasks but degrades significantly with complexity due to its directional exploration strategy and poor scalability. \textit{CEM} faces similar challenges, further limited by local optimization and lack of backtracking. While \textit{\mosaic{}-Roadmap} and \textit{Incremental \mosaic{}-Roadmap} show benefits of flexible exploration by removing sequential dependencies, their varying success rates across scenarios highlight the limitations of fixed strategies. These results demonstrate that \mosaic{}'s adaptive oracle-guided exploration enables more robust and efficient planning by dynamically adjusting its search strategy, enabling it to outperform baselines in complex scenarios. Our hardware validation tests (Fig. \ref{fig:experimental-setup}) trended similarly; View videos in our supplementary materials.

%% file: tex/conclusion.tex
\section{Conclusion}
\label{sec:conclusion}

We present \mosaic{}, a skill-centric framework for solving long-horizon manipulation tasks through the composition of generic, imperfect skills, and physics simulation. Our work makes three key contributions: First, we introduce a different approach to skill-based planning, where skills actively guide the planning process toward regions where they are likely to succeed, rather than being composed via strictly sequential, either forward or backward, chaining.
Second, we demonstrate that \mosaic{}'s modular architecture, consisting of \textit{generators} for producing local trajectories and \textit{connectors} for linking them through boundary value problems, with the use of physics simulation, allows us to avoid relying on hand-coded definitions of skill preconditions and effects.
Third, through experiments across multiple tasks, we show that \mosaic{} consistently achieves high success rates while reducing planning time compared to existing approaches, demonstrating that its adaptive oracle-guided exploration represents a fundamental advance in robust long-horizon manipulation planning. 
Looking forward, promising research directions include developing more sophisticated oracle modules, integrating foundation models for high-level reasoning, extending the framework to handle partial observability, and better utilizing the parallelization capabilities of physics simulators. We believe \mosaic{} offers a new paradigm for skill-centric planning that will advance progress toward more capable, adaptable robotic manipulation in unstructured environments.

%% file: tex/appendix.tex
\section{Appendix}
\label{sec:appendix}
In this appendix we will provide additional details about various practical components that we presented in this work. We include specifics about the oracle module used and the skills provided within the skill library. 

\subsection{Oracle}
\label{sec:appx:oracle}
The oracle module within the \mosaic{} algorithmic framework orchestrates the exploration of the skill trajectory space $\Tau$ by selecting skills to invoke and nodes to connect. Here we detail the specifics of the \textit{domain-independent statistical oracle} module that we used in the experiments in this work.

\subsubsection{Skill Selection}
The oracle chooses skills to invoke in two stages. First (potentially) restricting the library of skills to a subset and then choosing a particular skill within it.

\textbf{Skill Type Selection}: To decide whether to invoke a generator skill (to add nodes to the mosaic graph) or a connector skill (to increase the connectivity of the graph), the oracle uses the ratio of nodes to edges within the mosaic graph as a heuristic. Let $\mathcal{V}(\cdot)$ denote the number of nodes ($N$) or edges ($E$) in the mosaic graph, and let $p_{lb}$ and $p_{ub}$ be the lower and upper bounds to be used as thresholds to allow randomness in the skill type selection process. In addition, we define a random variable $T \sim \textit{U}(0, 1)$ to follow uniform distribution in the range of $0$ to $1$. Then, the type of skill set to activate is determined by the function:

\begin{equation*}
f_{\sigma}^{\text{type}}(N, E) = \begin{cases}
             \mathcal{C} & \text{if } P(T > \min (p_{up}, \max (\frac{\mathcal{V}(E)}{\mathcal{V}(N)}, p_{lb})))\\
             \Sigma & \text{else} 
       \end{cases} 
\end{equation*}

\noindent Informally, the oracle restricts the available skills to be only the connector skills if there are fewer nodes than vertices in the mosaic graph. The probability of this restriction increases with the ratio between nodes and edges.

\textbf{Individual Skill Selection:}
The oracle maintains statistics on the success rates of each skill, tracking the number of successful trajectories generated and their contributions to the overall connectivity of the mosaic graph. This allows the oracle to prioritize skills with higher success rates, promoting exploitation. Additionally, to encourage exploration, the oracle keeps track of how often each skill has been invoked. Less frequently used skills are given a higher likelihood of being selected.

Let $\alpha$ be a scalar parameter, $s_{\sigma}$ be the success rate of skill $\sigma$ and $t_{\sigma}$ be the number of times skill $\sigma$ was invoked. We define the following function for assigning noised priority for each skill option $\sigma_i \in f_{\sigma}^{\text{type}}$

\[f_{\sigma_i}^{\text{assign}}(s_{\sigma_i}, t_{\sigma_i}) = \alpha s_{\sigma_i} + (1-\alpha) \sqrt{\ln\frac{\sum_j (t_{\sigma_j} + 1)}{t_{\sigma_i} + 1}} + n \]
\[\text{where}\quad n \sim \mathcal{N}(0, 1)\]

Now, we pick the skill by choosing the one that has a maximum $f_{\sigma_i}^{\text{assign}}$ value

\[\sigma = \arg\max_{\sigma_i \in f_{\sigma}^{\text{type}}} f_{\sigma_i}^{\text{assign}}(s_{\sigma_i}, t_{\sigma_i}) \]

Once a skill $\sigma$ is selected, its associated parameters $\theta$ are sampled randomly from a predefined parameter distribution specific to that skill.

\subsubsection{Node Pair Selection for Connection}

Another decision made by the oracle module is deciding which pairs of nodes in the mosaic graph are worthwhile to attempt connecting. Since the mosaic graph is initially comprised of disconnected components representing different regions in the trajectory space where skills are effective, the choice of node-pairs to connect is critical for capitalizing on the recovered skill data toward solving the task at hand. In our oracle module, the selection of node-pairs is done in one of four randomization modes:
\begin{enumerate}
    \item Random selection: selecting a random node and connecting it to its nearest neighbor.
    \item Goal bias: if any nodes in the mosaic graph are known to be connected to a terminal node (i.e., a node whose trajectory terminates at a state satisfying the goal condition), choose one at random and attempt to connect it to its nearest neighbor.
    \item Start bias: among all nodes in the mosaic graph that are connected to the start node (i.e., the node holding the start state singleton trajectory), choose one at random and attempt connecting it to its nearest neighbor.
    \item Unification bias: if nodes connected to both the start and a terminal node exist, select a pair of these for a connection attempt.
\end{enumerate}

\textbf{A note on node pair selection.} To avoid repeated unsuccessful connection attempts, the oracle assigns penalties to node pairs with repeated failed connections by increasing the distance between them according to the number of failed connection attempts made. This discourages the selection of these nodes for future connections by modifying the results of nearest neighbor search, thereby guiding the search toward more promising regions of the mosaic graph. Additionally, if two nodes are already connected, they are not considered as pair candidates. 

Let $r \sim \textit{U}(0, 1)$ and let $p_{s}, p_{g}, p_{s-g}$ be the cutoffs for determining the mode of selection. Additionally, let $\textsc{modes} = \{\textsc{start}, \textsc{goal}, \textsc{start-goal}, \textsc{random}\}$ be the set of possible selection modes. Then, the mode selection function is defined as follows:

\begin{equation*}
m(\textsc{modes}) = \begin{cases}
                \textsc{start} \quad \text{if } r < p_{s} \\
                \textsc{goal} \quad \text{if } p_{s} < r < p_{g} \\
                \textsc{start-goal} \quad \text{if } p_{g} < r < p_{s-g} \\
                \textsc{random} \quad \text{if } r > p_{s-g} \\
       \end{cases} 
\end{equation*}

\subsection{Skills}
\label{sec:appx:skills}
This section provides detailed explanations of the skills we used throughout our experiments, complementing their introduction in Sec. \ref{sec:experimental_analysis}.

\subsubsection{\textit{Push} Skill}
In this work, we define two types of \textit{Push} skills: generator and connector. The generator push skill uses a \textit{diffusion} push policy to generate random object pushes on the tabletop. Its parameters include the object's start and goal poses, as well as a seed (within a predefined range) for the diffusion inference process. 
The connector push skill includes two modules: the push diffusion policy for connecting two object poses (corresponding to the end and start of the trajectories within two nodes in the mosaic graph), and the other is a motion planner for moving the robot to the beginning of the push motion and for matching the robot configuration in the second node after performing the push action. The only parameter for the connector skill is the seed.

\textbf{Additional details about Diffusion Policy:} To implement the push skill we use Diffusion Policy  \cite{chi2024diffusionpolicy}. These types of policies are generative models that learn a denoising process to recover dynamically-feasible trajectories from noise \citep{janner2022diffuser, carvalho2023mpd, shaoul2024multirobotmotionplanningdiffusion}. Given a dataset of trajectories, these score-based models aim to generate new trajectories following the underlying data distribution, conditioned on task objectives like start and termination conditions.
The diffusion inference process involves a $K$-step denoising process that transforms a noisy trajectory $^{K}\ve{\tau}_i$ to a feasible trajectory $^{0}\ve{\tau}_i$. Using Langevin dynamics sampling, at each step $k \in {K,\dots, 1}$, a trajectory-space mean $\mu^i_{k-1}$ is sampled from the network $\mu_\theta$:
\begin{equation*}
    \mu^i_{k-1} = \mu_\theta(^k\ve{\tau}_i)
\end{equation*}
The next trajectory is then sampled with a variance schedule, incorporating potential guidance:
\begin{equation*}
    ^{k-1}\ve{\tau}_i \sim \mathcal{N} \big(\mu^i_{k-1} + \underbrace{\eta \beta_{k-1} \nabla_{\ve{\tau}} \mathcal{J}(\mu_{k-1}^{i})}_{\text{Guidance}}, \beta_{k-1} \big)
\end{equation*}
Here, $\nabla_{\ve{\tau}} \mathcal{J}(\mu_{k-1}^{i})$ represents the gradient of additional trajectory-space objectives. The diffusion policy incorporates a receding horizon control, where, in this work, a simulator was used to update both the robot's state and the environment's state as feedback. For further details on diffusion policies, please refer to \citet{chi2024diffusionpolicy}. To train our Push skill, we collected 10,000 examples of pushing by randomly sampling start and goal poses of an object. For each example, we computed the object's axis-aligned bounding box (AABB) and push direction, then determined the initial push pose based on these factors. Closed-loop control (Jacobian control) was used to execute the push and bring the object to its goal pose. A diffusion policy network was trained on this dataset, where the observations include the robot's configuration, end-effector pose, object pose, and object goal pose. The network's output was joint actions to be executed on the robot.

\subsubsection{\textit{Pick} Skill, Score-based Geometric Antipodal Grasp}
The pick skill's policy begins with selecting an antipodal grasp that maximizes a score function while ensuring a valid inverse kinematics (IK) solution, followed by a screw-based motion planner that moves the robot through a sequence of poses -- pre-grasp, grasp, and retract. The procedure for selecting an antipodal grasp is composed of several stages:

\textbf{Point Cloud Acquisition and Surface Normal Estimation.}
A dense point cloud of the object is obtained, typically from an RGB-D sensor. Surface normals are estimated for each point by identifying the $k$-nearest neighbors (e.g., $k=30$) and fitting a local plane to these neighbors via Principal Component Analysis (PCA), where the normal is defined as the eigenvector corresponding to the smallest eigenvalue. 

\textbf{Sampling Antipodal Grasp Candidates.}
From the processed point cloud, a set of antipodal grasp candidates is generated based on the object's axis-aligned bounding box (AABB). Each candidate defines a potential gripper pose relative to the object and a closing direction corresponding to the line connecting the gripper fingers.

\textbf{Cropping Surface Normals and Evaluating Candidates.}
For each grasp candidate, the surface normals within the region between the hypothetical fingers are extracted. A grasp quality score is assigned based on the alignment of these normals with the gripper line, calculated as:
\begin{equation}
    \text{Score} = \frac{1}{N} \sum_{i=1}^{N} \left| n_i \cdot d_g \right|,
\end{equation}
where $n_i$ denotes a surface normal in the grasp region and $d_g$ is the direction vector between the fingers.

\textbf{Inverse Kinematics (IK) Verification.}
Candidates are ranked by their computed scores and checked for IK feasibility. This ensures that the selected grasp can be achieved without violating joint limits or causing collisions. The first candidate that satisfies the IK constraint is selected for execution.

The motion is decomposed into a sequence of three key gripper poses. The \textit{pre-grasp pose} is positioned slightly offset from the grasp pose along the approach direction, allowing the end-effector to approach the object safely. The \textit{grasp pose} is the target configuration determined during grasp selection. Finally, the \textit{retract pose} moves the gripper backward along the same approach direction after the object has been securely grasped. 

To generate smooth transitions between these poses, we employ a \textit{screw-based motion planner}.

\textbf{Screw Motion.}
To ensure continuous and physically plausible trajectories, we use \textit{screw motion} interpolation, which, based on Chasles' theorem, expresses the most general rigid body displacement as a rotation about and translation along a common axis, referred to as the \textit{screw axis}.

Let the start and goal gripper poses be represented by the rigid body transformations $A_{\text{start}}$ and $A_{\text{goal}}$ in $\text{SE}(3)$. The relative transformation between them is given by $A_{\text{rel}} = A_{\text{start}}^{-1} A_{\text{goal}}$. From this, we extract a twist $\xi$, consisting of an angular velocity vector $\omega \in \mathbb{R}^3$ and a linear velocity vector $v \in \mathbb{R}^3$. The interpolated motion at a normalized time parameter $s \in [0, 1]$ is then defined by:
\[
A(s) = A_{\text{start}} \exp\left( s \, \hat{\xi} \right),
\]
where $\hat{\xi} \in \mathfrak{se}(3)$ denotes the twist in matrix form, and $\exp$ represents the matrix exponential.

As $s$ progresses from $0$ to $1$, the end-effector traces a helical path that combines rotation and translation along the screw axis. This interpolation leads to smooth and continuous motion, avoiding abrupt changes in orientation or velocity, and is particularly effective for precise tasks such as transitioning from pre-grasp to grasp.

To execute the screw motion on an articulated manipulator, the desired end-effector twist $\xi$ must be mapped to joint-space velocities. This is accomplished through the robot's Jacobian matrix $J(q)$, which relates joint velocities $\dot{q}$ to the end-effector's spatial velocity. At each interpolation step, we solve for the joint velocities using:
\[
\dot{q} = J(q)^\dagger \, \xi,
\]
where $J(q)^\dagger$ denotes the (possibly damped) pseudoinverse of the Jacobian at the current configuration $q$. The joint positions are then integrated over time to produce a full joint-space trajectory that realizes the screw motion in task space.

When the Pick skill was used as a \textit{generator}, its parameters included the object's pose and a seed (within a predefined range) for the probabilistic components of the grasp selection process. When used as a \textit{connector}, the only parameter was the seed.

\subsection{Baselines: Additional Details}
We describe the algorithmic details of the baselines used in our evaluation. The first two baselines, \textit{Skills as Options} and \textit{CEM}, employ more directed strategies that sequentially compose skills toward the goal. In contrast, the \textit{\mosaic{}-Roadmap} variants explore the trajectory space more freely by constructing a roadmap of skill trajectories and searching over it. 

The \textbf{Skills as Options} algorithm mimics the behavior of skill chaining by exploring the space of skills incrementally, with each new skill beginning at the termination state of another. To realize this behavior, we create an action space $\hat{\mathcal{A}}$ comprised of modified generator skills that operate under start-state conditioning, effectively solving an initial-value problem when generating skill trajectories. Specifically, the action space $\hat{\mathcal{A}} := \hat {\mathcal{G}}$ with each skill being a mapping of the form $\hat {\mathcal{G}}_i: \mathcal{X} \times \Theta_{\mathcal{G}_i} \rightarrow \Tau$. The \textit{Skills as Options} algorithm searches over the space of world states $x \in \mathcal{X}$ in a breadth-first fashion. The algorithm begins by creating a node for the initial state $x_{\text{start}}$ and generates successor nodes via the skills in $\hat{\mathcal{A}}$ such that each successor set to a node $x$ is  $\{x' \mid x' = \ve{\tau}'(1), \ve{\tau}' = \hat {\mathcal{G}}_i(x, \theta), \theta \in \Theta_{\hat{\mathcal{G}}_i} \}$. Skill parameters are randomly chosen. The motion planner skill is goal-conditioned and attempts to connect a state $x$ to a goal state. The search terminates when a node has an associated state $x$ where $\xi_\text{goal}(x) = 1$. Finally, a path is reconstructed and returned.

\textbf{CEM} (Cross-Entropy Method \cite{cem}) approaches long-horizon planning by identifying promising sequences of skill parameters in a receding horizon manner. At each iteration, it samples a batch of skill-parameter sequences from $\hat{\mathcal{A}}$, evaluates them based on task progress, selects the top-performing candidates, and refines the sampling distribution accordingly. It then executes the first skill in the most promising sequence, updates the world state, and repeats this process until reaching the goal or exhausting the allowed iterations.

\textbf{\mosaic{}-Roadmap} deviates from the sequential nature of \textit{Skills as Options}, which requires discovered skills to begin at termination states of other skills, by allowing arbitrary generation of skill trajectories without conditioning on any previously discovered skill motions. This baseline is inspired by the Probabilistic Roadmap (PRM) \cite{PRM} structure and is modifying the \textit{oracle} in \mosaic{} accordingly. Specifically, \mosaic{}-Roadmap follows a two-phase approach: roadmap generation and path computation. To generate the roadmap, the algorithm repeatedly uses generator skills $\mathcal{G}_i \in \mathcal{A}$ to obtain skill trajectories $\ve{\tau}$ and attempts to connect their start $\ve{\tau}(0)$ and goal $\ve{\tau}(1)$ to the $k$ nearest generated trajectories $\ve{\tau}'$ in the roadmap with all connector skills $\mathcal{C}_i \in \mathcal{A}$ (In this work the distance metric between world state $x, x' \in \mathcal{X}$ is the square root of the sum of squared per-object pose differences (position difference and quaternion distance).
Once the roadmap is constructed, two vertices, $\{x_\text{start}\}$ and $\{x_\text{goal}\}$—each representing a singleton skill trajectory corresponding to the start and goal configurations—are added to the roadmap. The algorithm then attempts to connect these vertices to their nearest neighbors using the available connector skills $\mathcal{C}_j \in \mathcal{A}$. If successful, Dijkstra's algorithm \cite{dijkstra1959note} is used to compute a sequence of skill trajectories that connect them. If no such sequence is found, the algorithm reports failure.

\textbf{Incremental \mosaic{}-Roadmap} is our closest baseline to \mosaic{}. Its operation is generally similar to \mosaic{}-Roadmap, with the main difference between them being that the oracle for this baseline calls for additional rounds of skill generation and connection as long as the start and goal nodes are in disconnected components on the roadmap. This ensures that the algorithm will keep progressing towards a solution as long as one does not yet exist within the roadmap. 